\documentclass{article}
\usepackage[final]{dlde_neurips_2022}
\usepackage[utf8]{inputenc} % allow utf-8 input
\usepackage[T1]{fontenc}    % use 8-bit T1 fonts
\usepackage{hyperref}       % hyperlinks
\usepackage{url}            % simple URL typesetting
\usepackage{booktabs}       % professional-quality tables
\usepackage{amsfonts, amsmath}       % blackboard math symbols
\usepackage{nicefrac}       % compact symbols for 1/2, etc.
\usepackage{microtype}      % microtypography
\usepackage{tabularx}
\usepackage{multirow}

\usepackage{graphicx}
\usepackage{subfigure}
\usepackage{caption}
\usepackage{xcolor}

\usepackage{appendix}
\usepackage{fancyvrb}

\usepackage{amssymb}% http://ctan.org/pkg/amssymb
\usepackage{pifont}% http://ctan.org/pkg/pifont
\newcommand{\cmark}{\ding{51}}%
\newcommand{\xmark}{\ding{55}}%

\usepackage{wrapfig}

\title{A Neural ODE Interpretation of Transformer Layers}
\author{%
    Yaofeng Desmond Zhong,$\quad$ Tongtao Zhang,$\quad$ Amit Chakraborty,$\quad$ Biswadip Dey
    \\
    Siemens Technology, Princeton, NJ 08536, USA.
    \\
    \texttt{\{yaofeng.zhong, tongtao.zhang, amit.chakraborty, biswadip.dey\}@siemens.com}
}
\begin{document}
\maketitle
%
%
%%%%%%%%%%%%%%%%%%%%%%%%%%%%%%%%%%%%%%%%%%%%%%
\begin{abstract}
Transformer layers, which use an alternating pattern of multi-head attention and multi-layer perceptron (MLP) layers, provide an effective tool for a variety of machine learning problems. As the transformer layers use residual connections to avoid the problem of vanishing gradients, they can be viewed as the numerical integration of a differential equation. In this extended abstract, we build upon this connection and propose a modification of the internal architecture of a transformer layer. The proposed model places the multi-head attention sublayer and the MLP sublayer parallel to each other. Our experiments show that this simple modification improves the performance of transformer networks in multiple tasks. Moreover, for the image classification task, we show that using neural ODE solvers with a sophisticated integration scheme further improves performance.
\end{abstract}
%%%%%%%%%%%%%%%%%%%%%%%%%%%%%%%%%%%%%%%%%%%%%%
%
%
%
%%%%%%%%%%%%%%%%%%%%%%%%%%%%%%%%%%%%%%%%%%%%%%
%%%%%%%%%%%%%%%%%%%%%%%%%%%%%%%%%%%%%%%%%%%%%%
\section{Introduction}
%%%%%%%%%%%%%%%%%%%%%%%%%%%%%%%%%%%%%%%%%%%%%%
%%%%%%%%%%%%%%%%%%%%%%%%%%%%%%%%%%%%%%%%%%%%%%
%
Over the last few years, the transformer layer introduced by \cite{vaswani2017attention} has become a key component in deep learning models used in natural language processing \citep{devlin-etal-2019-bert, NEURIPS2020_1457c0d6, https://doi.org/10.48550/arxiv.1909.08053}, image and video processing \citep{dosovitskiy2021an, Arnab_2021_ICCV, 9716741}, and audio and speech processing \citep{8462506}. Current state-of-the-art techniques in language processing (e.g., machine translation, natural language understanding, and information/knowledge extraction) rely heavily on the use of transformer layers to encode information in its word vector about the relevant context of a given word. This allows the model to focus on relevant contexts at different length scales. Although the transformer layers were originally introduced as a sequence-to-sequence transduction model, they have also demonstrated superior performance in various computer vision tasks beyond image classification, e.g., semantic segmentation \citep{9578646, strudel2021segmenter, ding2022davit}, object detection \citep{carion2020end, song2022vidt}, and view synthesis \citep{kulhanek2022viewformer, lin2022visionnerf}.

Inside a multilayer transformer network, each transformer layer consists of a multi-head attention sublayer followed by an MLP sublayer, creating an alternating pattern of these sublayers throughout the network. Moreover, both of these sublayers use residual connections to avoid the vanishing gradient problem and facilitate the training of very deep transformer networks. Prior work \citep{Haber_2017, Haber_Ruthotto_Holtham_Jun_2018, pmlr-v80-lu18d} has shown that the forward propagation through residual connections can be viewed as Euler discretization of a time-varying ordinary differential equation (ODE). This insight suggests a connection between transformer networks and differential equations that can potentially be exploited to further improve the performance of transformer networks (e.g., higher accuracy, fewer parameters). Indeed, the residual connections alongside the alternating pattern of multi-head attention and MLP sublayers can be interpreted as numerical integration via the Lie-Trotter splitting scheme \citep{lu2020understanding, NEURIPS2021_2bd388f7}. In this work, we show that the connection between transformers and ODEs can be leveraged to design a new architecture wherein the multi-head attention and MLP sublayers are placed parallelly, not sequentially, inside the individual transformer layers. Our experiments show that the proposed model performs better than the original transformer layer when tested on image classification, machine translation, and language modeling tasks.

A growing body of work has focused on improving transformer networks by reorganizing the sublayers and leveraging the connection between transformers and ODEs. Macaron Net \citep{lu2020understanding} draws inspiration from numerical integration techniques to use a multi-head attention sublayer sandwiched between two MLP sublayers.  \cite{NEURIPS2021_2bd388f7} use a temporal evolution scheme to avoid the computationally expensive step of calculating dot-product attention at each transformer layer; instead, it computes the dot-product attention at the initial step and then time-evolves it through the layers. The proposed model is similar in spirit to this line of work - it modifies the internal architecture of the transformer layer and places the multi-head attention sublayer and the MLP sublayer side-by-side. On the other hand, \cite{press-etal-2020-improving} have explored the effect of reordering the individual sublayers and changing their numbers while keeping the total number of model parameters fixed. Their work shows that a transformer network can improve its performance by concentrating the multi-head attention and MLP layers in the lower and upper stages of the network, respectively; however, this performance improvement is not uniform across all tasks.

The main contributions of this work are as follows: \\
\indent $\quad \bullet$ By leveraging the connection between transformer layers and ODEs, we propose a novel variant of the transformer layer wherein the multi-head attention and MLP sublayers are placed side-by-side.\\
\indent $\quad \bullet$ Through numerical experiments, we demonstrate that the proposed model performs better than the original transformer layer across multiple tasks (image classification on CIFAR-100, machine translation on WMT-2014 English-German dataset, and language modeling on WikiText-103). Our experiments have been carried out with small models due to resource constraints.\\
\indent $\quad \bullet$ We also demonstrate that the performance of the proposed model can be further improved by using neural ODE solvers with sophisticated integration schemes (e.g., RK4).
%
%

%
%
%%%%%%%%%%%%%%%%%%%%%%%%%%%%%%%%%%%%%%%%%%%%%%
%%%%%%%%%%%%%%%%%%%%%%%%%%%%%%%%%%%%%%%%%%%%%
\section{Proposed Architecture}
%%%%%%%%%%%%%%%%%%%%%%%%%%%%%%%%%%%%%%%%%%%%%
%%%%%%%%%%%%%%%%%%%%%%%%%%%%%%%%%%%%%%%%%%%%%%
%
%
%
\begin{figure}[b!]
\vspace{-1em}
    \centering
    \subfigure[]{
        \centering
        \includegraphics[width=0.3\textwidth]{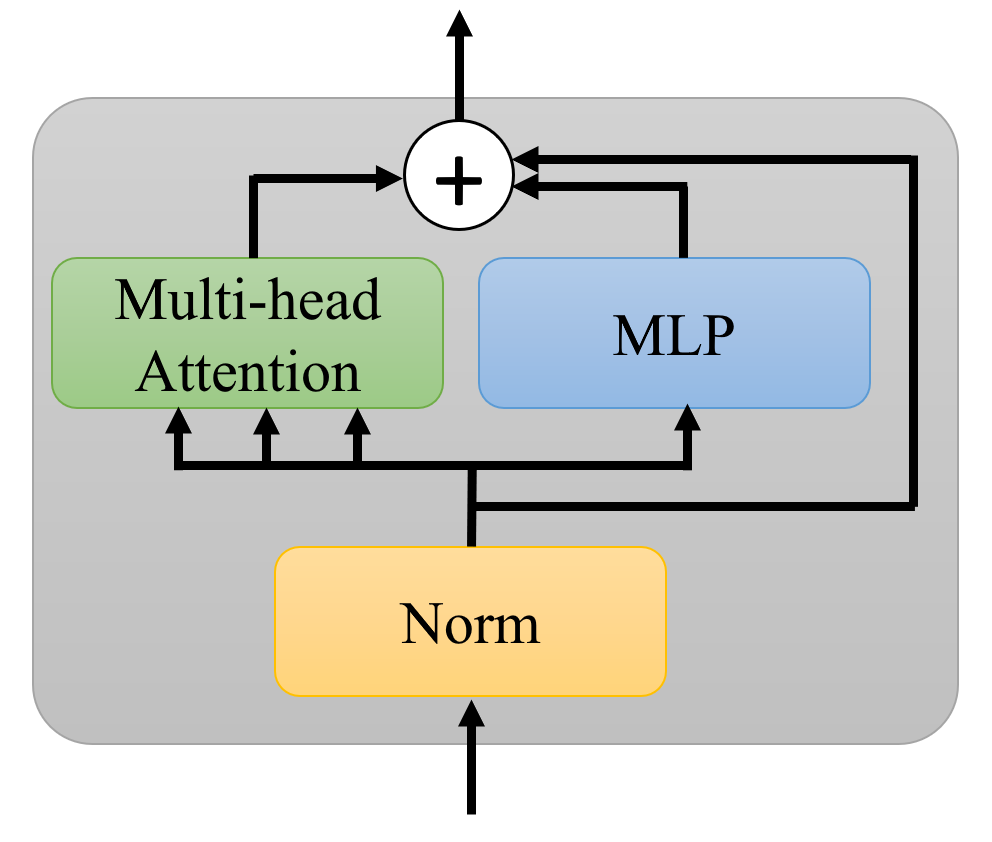}
        \label{fig:proposed}
    }
    \subfigure[]{
        \centering
        \includegraphics[width=0.15\textwidth]{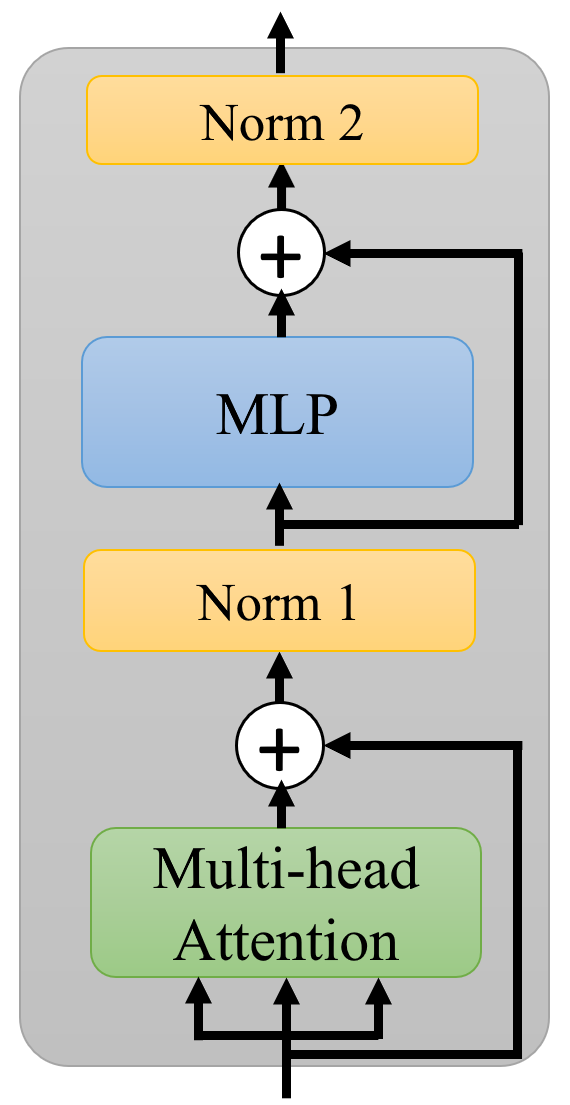}
        \label{fig:original}
    }
    \subfigure[]{
        \centering
        \includegraphics[width=0.15\textwidth]{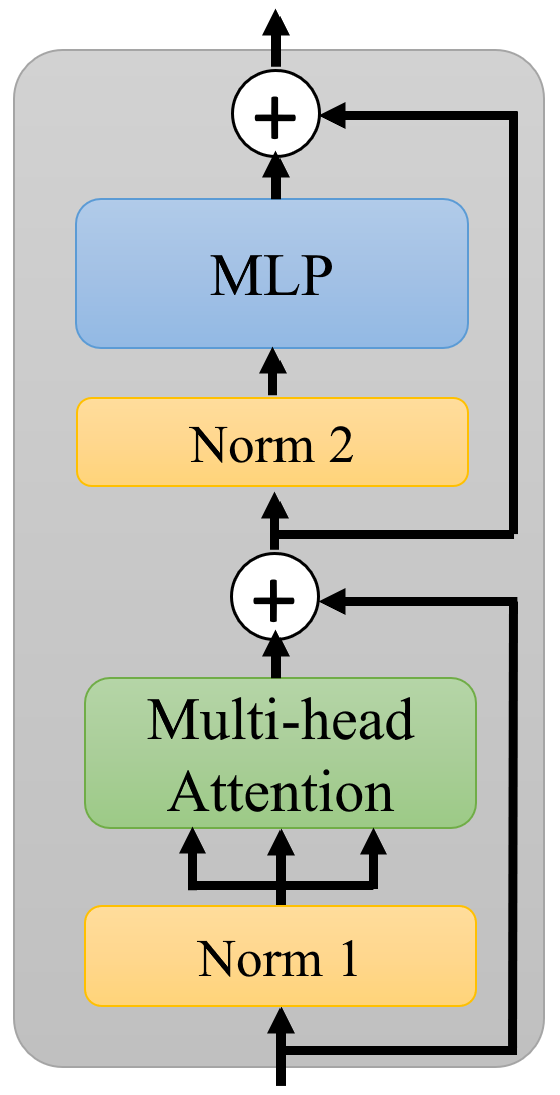}
        \label{fig:vit}
    }
    \vspace{-0.5em}
    \caption{\small{This figure shows the proposed model (left panel: a) along with the original version of the transformer layer (center panel: b) and the transformer layer used in the vision transformer (right panel: c).}}
    \label{fig:compare}
\end{figure}
By letting $X^m := [x_1^m, \; x_2^m, \; \cdots, \; x_L^m]$ denote the input to the $m$-th transformer layer, the operation carried out by the multi-head attention sublayer can be expressed as
\begin{equation}
    \hat{x}_i^m = x_i^m + G(x_i^m, X^m), \qquad 1 \leq i \leq L,
    \label{MSAlayer}
\end{equation}
where $L$ is the length of the input sequence and the function $G$ represents the multi-head dot-product attention. $\hat{x}_i^m$, i.e., the output from this sublayer, is then fed to the MLP sublayer and undergoes the following transformation to yield $X^{m+1} = [x_1^{m+1}, \; x_2^{m+1}, \; \cdots, \; x_L^{m+1}]$
\begin{equation}
    x_i^{m+1} = \hat{x}_i^m + F(\hat{x}_i^m), \qquad 1 \leq i \leq L,
    \label{FFNlayer}
\end{equation}
where the function $F$ represents the sequence of linear mappings and activation functions. As \cite{lu2020understanding} and \cite{NEURIPS2021_2bd388f7} have highlighted, \eqref{MSAlayer}-\eqref{FFNlayer} can viewed as the numerical integration (over the time interval $[m,m+1]$) of the following ODE via the Lie-Trotter splitting scheme
\begin{equation}
    \frac{dx_i}{dt} = F(x_i) + G(x_i,X),
    \label{ODEisALL}
\end{equation}
where $X := [x_1, \; x_2, \; \cdots, \; x_L]$.

This interpretation paves the way for multiple approaches to realize transformer layers. As neural ODE solvers and their variants \citep{chen2018neural, NEURIPS2020_dissectNeuralODE, pmlr-v139-kidger21a} provide a means to run backpropagation through any black-box ODE solver, a transformer layer can be implemented using neural ODE networks. Alternatively, the time integration of \eqref{ODEisALL} over the interval $[m,m+1]$ can also be approximated as
\begin{equation}
    x_i^{m+1} = x_i^{m} + \Big[ F(x_i^m) + G(x_i^m,X^m) \Big] , \qquad 1 \leq i \leq L.
    \label{PropModel}
\end{equation}
Our proposed model implements \eqref{PropModel} by placing the multi-head attention and MLP sublayers side-by-side (Figure~\ref{fig:compare}).
Moreover, if they share their weights, a $D$-layer deep stack of transformer layers can be viewed as the numerical integration of \eqref{ODEisALL} over the time interval $[0,D]$. This perspective offers a means to introduce weight sharing into a transformer network in a gradual way for understanding the trade-off between model performance and the number of model parameters. For example, a 12-layer transformer network can be replaced by a sequence of six 2-layer, weight sharing transformers (which can be implemented by either stacking two layers of the proposed model or via integration over a longer horizon using a neural ODE).
\begin{table}[b!]
\vspace{-2em}
\caption{\small{Performance on CIFAR-100 classification task of our proposed model with different levels of weight sharing. All the experiments are done without dropout and stochastic depth. \textbf{Top two rows}: 12 independent layers mean there's no weight sharing. the only difference between proposed architecture and DeiT-Ti is indicated in Figure \ref{fig:compare}; \textbf{Bottom row}: 1 independent layer means sharing weights for all 12 layers.}}
\label{tab:diff-ind-layers}
\vskip 0.05in
\centering
\begin{tabular}{c | c | c | c | c }
    \toprule[1.25pt]
    \textbf{Model} & \textbf{\# layers} & \textbf{\# independent layers} & \textbf{\# parameters} & \textbf{Top-1 accuracy} \\
    \midrule[0.75pt]
    DeiT-Ti & 12 & 12 & 5.5M & 66.02\% \\
    \midrule[0.75pt]
    \multirow{6}{*}{Proposed model} & 12 & 12 & 5.5M & 70.92\%\\
    & 12 & 6 & 2.9M & 67.42\%\\
    & 12 & 4 & 2.0M & 66.40\%\\
    & 12 & 3 & 1.5M & 63.68\%\\
    & 12 & 2 & 1.1M & 61.05\%\\
    & 12 & 1 & 0.7M & 53.33\%\\
    \bottomrule[1.25pt]
\end{tabular}
\end{table}
%
%

%
%
%%%%%%%%%%%%%%%%%%%%%%%%%%%%%%%%%%%%%%%%%%%%%%
%%%%%%%%%%%%%%%%%%%%%%%%%%%%%%%%%%%%%%%%%%%%%%
\section{Experiments}
%%%%%%%%%%%%%%%%%%%%%%%%%%%%%%%%%%%%%%%%%%%%%%
%%%%%%%%%%%%%%%%%%%%%%%%%%%%%%%%%%%%%%%%%%%%%%
%
In this section, we investigate the performance of the proposed model for both computer vision and language processing tasks. However, due to limited computing resources, we use smaller models for comparing the performance between the baseline transformer architecture and the proposed model. Therefore, the reported results of baselines may not be comparable to the state-of-the-art results.
%

%
%
%%%%%%%%%%%%%%%%%%%%%%%%%%%%%%%%%%%%%%%%%%%%%%
%%%%%%%%%%%%%%%%%%%%%%%%%%%%%%%%%%%%%%%%%%%%%%
\subsection{Image Classification}
%%%%%%%%%%%%%%%%%%%%%%%%%%%%%%%%%%%%%%%%%%%%%%
%%%%%%%%%%%%%%%%%%%%%%%%%%%%%%%%%%%%%%%%%%%%%%
%
First, we investigate the proposed model's performance in image classification tasks. In particular, we take the DeiT-Ti model proposed by \cite{touvron2020training} and modify the transformer layer architecture to implement our proposed model. Then, to compare the performance of DeiT-Ti and variants of our proposed models on CIFAR-100, we use top-1 accuracy as the metric.

Table~\ref{tab:diff-ind-layers} demonstrates the performance of the side-by-side sublayers with different levels of weight sharing. We observe that by only changing the attention block and MLP block from sequential to parallel, the top-1 accuracy increases from 66.02\% to 70.92\%. This architecture does not use any weight sharing. Increasing the amount of weight sharing reduces the number of independent layers and results in fewer trainable parameters. However, as the number of parameters decreases, the model performance also deteriorates. 

By leveraging the connection between differential equations and the transformer layers, we also replace the Euler scheme with a more sophisticated Runge-Kutta (RK4) integration scheme. This further increases the top-1 accuracy from 70.92\% (Euler) to 72.66\% (RK4).
\begin{figure}[t!]
\vspace{-1em}
    \centering
    \subfigure[]{
        \centering
        \includegraphics[width=0.275\textwidth]{figures/arch_proposed.png}
        \label{fig:proposed}
    }    
    \subfigure[]{
        \centering
        \includegraphics[width=0.275\textwidth]{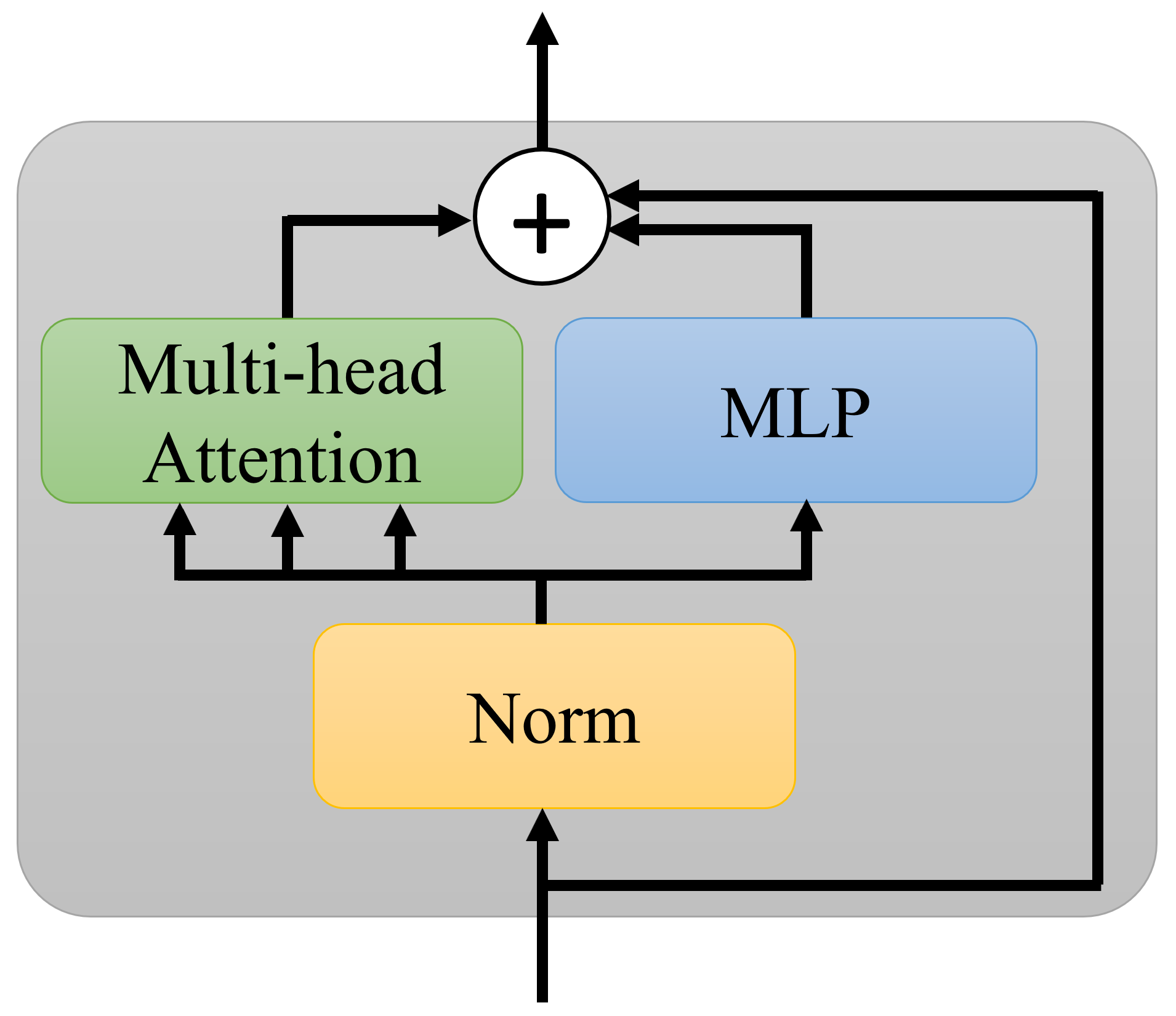}
        \label{fig:original}
    }
    \subfigure[]{
        \centering
        \includegraphics[width=0.275\textwidth]{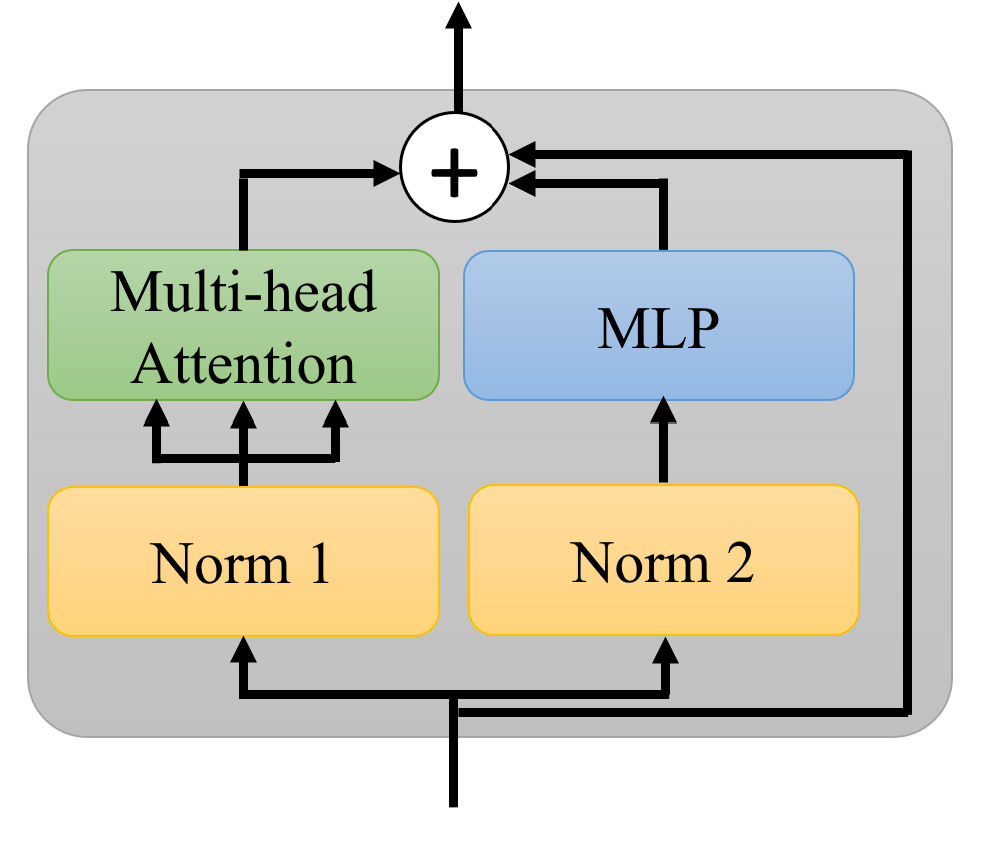}
        \label{fig:vit}
    }
    \vspace{-1em}
    \caption{\small{Variations of normalization implementations in the proposed model. In the image classification task, they lead to different top-1 accuracy: (a) 70.92\%; (b) 67.42\%; (c) 67.03\%, all of which beat DeiT-Ti (66.02\%).}}
    \label{fig:variation}
\end{figure}

Furthermore, we study the effects of dropout \citep{srivastava2014dropout} and stochastic depth \citep{huang2016deep} in Table~\ref{tab:dropout}. We conclude that neither dropout nor stochastic depth is helpful in the proposed architecture. 
\begin{table}[h]
\vspace{-1.5em}
\caption{\small{Performance on CIFAR-100 classification task of our proposed model. Neither dropout nor stochastic depth benefits training for our proposed model.}}
\label{tab:dropout}
\vskip 0.05in
\centering
\begin{tabular}{c | c | c | c}
    \toprule[1.25pt]
    Dropout & Stochastic depth & Top-1 accuracy (12 ind. layers) & Top-1 accuracy (1 ind. layers)   \\
    \midrule[1pt]
    \cmark & \xmark & 59.81\% & 40.71\%\\
    \xmark & \cmark & 71.49\% & 27.57\%\\
    \xmark & \xmark & 70.92\% & 53.33\%\\
    \bottomrule[1.25pt]
\end{tabular}
\vspace{-1.5em}
\end{table}

We also conducted a small ablation study on the placement of the normalization layer in our proposed model. Figure~\ref{fig:variation} shows that all three variations perform better than the original DeiT-Ti model. Moreover, our proposed normalization approach (Figure~\ref{fig:variation}a) yields the best performance.
%
%

%
%
%%%%%%%%%%%%%%%%%%%%%%%%%%%%%%%%%%%%%%%%%%%%%%
%%%%%%%%%%%%%%%%%%%%%%%%%%%%%%%%%%%%%%%%%%%%%%
\subsection{Natural Language Processing}
%%%%%%%%%%%%%%%%%%%%%%%%%%%%%%%%%%%%%%%%%%%%%%
%%%%%%%%%%%%%%%%%%%%%%%%%%%%%%%%%%%%%%%%%%%%%%
%
We leverage the open source toolkit Fairseq \citep{ott2019fairseq} to investigate the performance of the proposed model in neural machine translation and language modeling. Additional details about the size of the architectures used in this section are provided in the Appendix. 
%
%

%
%
%%%%%%%%%%%%%%%%%%%%%%%%%%%%%%%%%%%%%%%%%%%%%%
%%%%%%%%%%%%%%%%%%%%%%%%%%%%%%%%%%%%%%%%%%%%%%
\subsubsection{Neural Machine Translation}
%%%%%%%%%%%%%%%%%%%%%%%%%%%%%%%%%%%%%%%%%%%%%%
%%%%%%%%%%%%%%%%%%%%%%%%%%%%%%%%%%%%%%%%%%%%%%
%
%
\begin{wraptable}[6]{r}{0.59\textwidth}
\vspace{-3.75em}
\caption{\small{Performance on Neural Machine Translation}}
\label{tab:translation}
\vspace{-0.75em}
% \begin{small}
\centering
\begin{tabular}{c | c | c | c}
    % \toprule[1.25pt]
    % Dropout & Model & Validation loss & BLEU score   \\
    % \midrule[1pt]
    % \multirow{2}{*}{0.1} & sequential & 3.455 & 18.7\\
    %  & parallel & 3.519 & 18.5 \\
    % \midrule[0.5pt]
    % \multirow{2}{*}{0.0} & sequential & 3.788 & 17.1\\
    %  & parallel & 3.626 & 17.9\\
    % \bottomrule[1.25pt]
    \toprule[1.25pt]
    \textbf{Model} & 
    \textbf{Dropout} & \textbf{Validation loss} & \textbf{BLEU score}   \\
    \midrule[1pt]
    \multirow{2}{*}{Sequential} & 0.0 & 3.788 & 17.1\\
     & 0.1 & 3.455 & 18.7 \\
    \midrule[0.5pt]
    \multirow{2}{*}{Parallel} & 0.0 & 3.626 & 17.9\\
     & 0.1 & 3.519 & 18.5\\
    \bottomrule[1.25pt]
\end{tabular}
\end{wraptable}
To train small transformer models on the WMT-2014 English-German translation dataset, we follow the training procedure specified by \cite{ott-etal-2018-scaling, ott2019fairseq}. After training the models for 30 epochs, we compare their performance by computing the validation loss and detokenized BLEU score with SacreBLEU \citep{post2018call} as described by \cite{ott-etal-2018-scaling}. As shown in Table~\ref{tab:translation}, the proposed model outperforms the baseline.
%
%

%
%
%%%%%%%%%%%%%%%%%%%%%%%%%%%%%%%%%%%%%%%%%%%%%%
%%%%%%%%%%%%%%%%%%%%%%%%%%%%%%%%%%%%%%%%%%%%%%
\subsubsection{Neural Language Modeling}
%%%%%%%%%%%%%%%%%%%%%%%%%%%%%%%%%%%%%%%%%%%%%%
%%%%%%%%%%%%%%%%%%%%%%%%%%%%%%%%%%%%%%%%%%%%%%
%
%
\begin{wraptable}[8]{r}{0.4\textwidth}
\vspace{-3.5em}
\caption{\small{Performance on Neural Language Modeling}}
\label{tab:lm}
\vspace{-0.5em}
\centering
\begin{tabular}{c | c | c }
    % \toprule[1.25pt]
    % Model & \multicolumn{3}{c|}{Sequential} & \multicolumn{3}{c}{Parallel}   \\
    % \midrule[0.5pt]
    % Dropout & 0.0 & 0.1 & 0.3 & 0.0 & 0.1 & 0.3\\
    % \midrule[1.25pt]
    % Perplexity & 65.07 & 72.72 & 101.09 & 60.19 & 76.21 & 113.27 \\
    % \bottomrule[1.25pt]
    \toprule[1.25pt]
    \textbf{Model} & \textbf{Dropout} & \textbf{Perplexity}  \\
    \midrule[0.5pt]
    \multirow{3}{*}{Sequential} & 0.0 & 65.07\\
    & 0.1 & 72.72\\
    & 0.3 & 101.09\\
    \midrule[1.25pt]
    \multirow{3}{*}{Parallel} & 0.0 & 60.19\\
    & 0.1 & 76.21\\
    & 0.3 & 113.27\\
    \bottomrule[1.25pt]
\end{tabular}
% \end{small}
\end{wraptable}
We follow the training and evaluation procedure specified by \cite{baevski2018adaptive} and \cite{ott2019fairseq} to train a small language model on the WikiText-103 dataset. Table~\ref{tab:lm} reports the metric of perplexity for the baseline and proposed model for different dropout rates. We can notice that for these small models, dropout doesn't lead to improved performance for both the baseline and the proposed architecture. With zero dropout, our proposed architecture performs better than the baseline.  
%
%

%
%
%%%%%%%%%%%%%%%%%%%%%%%%%%%%%%%%%%%%%%%%%%%%%%
\section{Conclusion}
%%%%%%%%%%%%%%%%%%%%%%%%%%%%%%%%%%%%%%%%%%%%%%
%
This work has proposed a new variant of the transformer layer by leveraging its connection with ODEs and has shown that the proposed model outperforms the original transformer layer. Furthermore, as shown by our initial results from an RK4-based neural ODE solver, one can extend this work to investigate the potential of using a time-dependent neural ODE to implement transformer networks.
%
%

%%%%%%%%%%%%%%%%%%%%%%%%%%%%%%%%%%%%%%%%%%%%%
%%%%%%%%%%%%%%%%%%%%%%%%%%%%%%%%%%%%%%%%%%%%%
\newpage
\small
\bibliography{References}

\begin{thebibliography}{}

\bibitem[Arnab et~al., 2021]{Arnab_2021_ICCV}
Arnab, A., Dehghani, M., Heigold, G., Sun, C., Lu\v{c}i\'c, M., and Schmid, C.
  (2021).
\newblock {ViViT}: A video vision transformer.
\newblock In {\em IEEE/CVF International Conference on Computer Vision (ICCV)}.

\bibitem[Baevski and Auli, 2019]{baevski2018adaptive}
Baevski, A. and Auli, M. (2019).
\newblock Adaptive input representations for neural language modeling.
\newblock In {\em International Conference on Learning Representations}.

\bibitem[Brown et~al., 2020]{NEURIPS2020_1457c0d6}
Brown, T., Mann, B., Ryder, N., Subbiah, M., Kaplan, J.~D., Dhariwal, P.,
  Neelakantan, A., Shyam, P., Sastry, G., Askell, A., Agarwal, S.,
  Herbert-Voss, A., Krueger, G., Henighan, T., Child, R., Ramesh, A., Ziegler,
  D., Wu, J., Winter, C., Hesse, C., Chen, M., Sigler, E., Litwin, M., Gray,
  S., Chess, B., Clark, J., Berner, C., McCandlish, S., Radford, A., Sutskever,
  I., and Amodei, D. (2020).
\newblock Language models are few-shot learners.
\newblock In {\em Advances in Neural Information Processing Systems},
  volume~33, pages 1877--1901.

\bibitem[Carion et~al., 2020]{carion2020end}
Carion, N., Massa, F., Synnaeve, G., Usunier, N., Kirillov, A., and Zagoruyko,
  S. (2020).
\newblock End-to-end object detection with transformers.
\newblock In {\em European Conference on Computer Vision (ECCV)}.

\bibitem[Chen et~al., 2018]{chen2018neural}
Chen, R.~T., Rubanova, Y., Bettencourt, J., and Duvenaud, D.~K. (2018).
\newblock Neural ordinary differential equations.
\newblock In {\em Advances in Neural Information Processing Systems},
  volume~31, pages 6571--6583.

\bibitem[Devlin et~al., 2019]{devlin-etal-2019-bert}
Devlin, J., Chang, M.-W., Lee, K., and Toutanova, K. (2019).
\newblock {BERT}: Pre-training of deep bidirectional transformers for language
  understanding.
\newblock In {\em Proceedings of the 2019 Conference of the North {A}merican
  Chapter of the Association for Computational Linguistics: Human Language
  Technologies, Volume 1 (Long and Short Papers)}, pages 4171--4186.

\bibitem[Ding et~al., 2022]{ding2022davit}
Ding, M., Xiao, B., Codella, N., Luo, P., Wang, J., and Yuan, L. (2022).
\newblock {DaViT}: Dual attention vision transformer.
\newblock In {\em European Conference on Computer Vision (ECCV)}.

\bibitem[Dong et~al., 2018]{8462506}
Dong, L., Xu, S., and Xu, B. (2018).
\newblock Speech-transformer: A no-recurrence sequence-to-sequence model for
  speech recognition.
\newblock In {\em 2018 IEEE International Conference on Acoustics, Speech and
  Signal Processing (ICASSP)}, pages 5884--5888.

\bibitem[Dosovitskiy et~al., 2021]{dosovitskiy2021an}
Dosovitskiy, A., Beyer, L., Kolesnikov, A., Weissenborn, D., Zhai, X.,
  Unterthiner, T., Dehghani, M., Minderer, M., Heigold, G., Gelly, S.,
  Uszkoreit, J., and Houlsby, N. (2021).
\newblock An image is worth 16x16 words: Transformers for image recognition at
  scale.
\newblock In {\em International Conference on Learning Representations}.

\bibitem[Dutta et~al., 2021]{NEURIPS2021_2bd388f7}
Dutta, S., Gautam, T., Chakrabarti, S., and Chakraborty, T. (2021).
\newblock Redesigning the transformer architecture with insights from
  multi-particle dynamical systems.
\newblock In {\em Advances in Neural Information Processing Systems},
  volume~34, pages 5531--5544.

\bibitem[Haber and Ruthotto, 2017]{Haber_2017}
Haber, E. and Ruthotto, L. (2017).
\newblock Stable architectures for deep neural networks.
\newblock {\em Inverse Problems}, 34(1):014004.

\bibitem[Haber et~al., 2018]{Haber_Ruthotto_Holtham_Jun_2018}
Haber, E., Ruthotto, L., Holtham, E., and Jun, S.-H. (2018).
\newblock Learning across scales---{M}ultiscale methods for convolution neural
  networks.
\newblock {\em Proceedings of the AAAI Conference on Artificial Intelligence},
  32.

\bibitem[Han et~al., 2022]{9716741}
Han, K., Wang, Y., Chen, H., Chen, X., Guo, J., Liu, Z., Tang, Y., Xiao, A.,
  Xu, C., Xu, Y., Yang, Z., Zhang, Y., and Tao, D. (2022).
\newblock A survey on vision transformer.
\newblock {\em IEEE Transactions on Pattern Analysis and Machine Intelligence}.

\bibitem[Huang et~al., 2016]{huang2016deep}
Huang, G., Sun, Y., Liu, Z., Sedra, D., and Weinberger, K.~Q. (2016).
\newblock Deep networks with stochastic depth.
\newblock In {\em European conference on computer vision}. Springer.

\bibitem[Kidger et~al., 2021]{pmlr-v139-kidger21a}
Kidger, P., Chen, R. T.~Q., and Lyons, T.~J. (2021).
\newblock "{Hey, that’s not an ODE}": Faster {ODE} adjoints via seminorms.
\newblock In {\em Proceedings of the 38th International Conference on Machine
  Learning}, pages 5443--5452.

\bibitem[Kulh{\'a}nek et~al., 2022]{kulhanek2022viewformer}
Kulh{\'a}nek, J., Derner, E., Sattler, T., and Babu{\v{s}}ka, R. (2022).
\newblock Viewformer: Nerf-free neural rendering from few images using
  transformers.
\newblock In {\em European Conference on Computer Vision (ECCV)}.

\bibitem[Lin et~al., 2022]{lin2022visionnerf}
Lin, K.-E., Yen-Chen, L., Lai, W.-S., Lin, T.-Y., Shih, Y.-C., and Ramamoorthi,
  R. (2022).
\newblock Vision transformer for nerf-based view synthesis from a single input
  image.
\newblock {\em arXiv:2207.05736}.

\bibitem[Lu et~al., 2020]{lu2020understanding}
Lu, Y., Li, Z., He, D., Sun, Z., Dong, B., Qin, T., Wang, L., and Liu, T.-y.
  (2020).
\newblock Understanding and improving transformer from a multi-particle dynamic
  system point of view.
\newblock In {\em ICLR 2020 Workshop on Integration of Deep Neural Models and
  Differential Equations}.

\bibitem[Lu et~al., 2018]{pmlr-v80-lu18d}
Lu, Y., Zhong, A., Li, Q., and Dong, B. (2018).
\newblock Beyond finite layer neural networks: Bridging deep architectures and
  numerical differential equations.
\newblock In {\em Proceedings of the 35th International Conference on Machine
  Learning}, pages 3276--3285.

\bibitem[Massaroli et~al., 2020]{NEURIPS2020_dissectNeuralODE}
Massaroli, S., Poli, M., Park, J., Yamashita, A., and Asama, H. (2020).
\newblock {Dissecting Neural ODEs}.
\newblock In {\em Advances in Neural Information Processing Systems},
  volume~33.

\bibitem[Ott et~al., 2019]{ott2019fairseq}
Ott, M., Edunov, S., Baevski, A., Fan, A., Gross, S., Ng, N., Grangier, D., and
  Auli, M. (2019).
\newblock fairseq: A fast, extensible toolkit for sequence modeling.
\newblock In {\em Proceedings of NAACL-HLT 2019: Demonstrations}.

\bibitem[Ott et~al., 2018]{ott-etal-2018-scaling}
Ott, M., Edunov, S., Grangier, D., and Auli, M. (2018).
\newblock Scaling neural machine translation.
\newblock In {\em Proceedings of the Third Conference on Machine Translation:
  Research Papers}, pages 1--9, Brussels, Belgium.

\bibitem[Post, 2018]{post2018call}
Post, M. (2018).
\newblock A call for clarity in reporting {BLEU} scores.
\newblock {\em arXiv:1804.08771}.

\bibitem[Press et~al., 2020]{press-etal-2020-improving}
Press, O., Smith, N.~A., and Levy, O. (2020).
\newblock Improving transformer models by reordering their sublayers.
\newblock In {\em Proceedings of the 58th Annual Meeting of the Association for
  Computational Linguistics}, pages 2996--3005.

\bibitem[Shoeybi et~al., 2019]{https://doi.org/10.48550/arxiv.1909.08053}
Shoeybi, M., Patwary, M., Puri, R., LeGresley, P., Casper, J., and Catanzaro,
  B. (2019).
\newblock {Megatron-LM}: Training multi-billion parameter language models using
  model parallelism.

\bibitem[Song et~al., 2022]{song2022vidt}
Song, H., Sun, D., Chun, S., Jampani, V., Han, D., Heo, B., Kim, W., and Yang,
  M.-H. (2022).
\newblock Vi{DT}: An efficient and effective fully transformer-based object
  detector.
\newblock In {\em International Conference on Learning Representations}.

\bibitem[Srivastava et~al., 2014]{srivastava2014dropout}
Srivastava, N., Hinton, G., Krizhevsky, A., Sutskever, I., and Salakhutdinov,
  R. (2014).
\newblock Dropout: a simple way to prevent neural networks from overfitting.
\newblock {\em The journal of machine learning research}, 15(1):1929--1958.

\bibitem[Strudel et~al., 2021]{strudel2021segmenter}
Strudel, R., Garcia, R., Laptev, I., and Schmid, C. (2021).
\newblock Segmenter: Transformer for semantic segmentation.
\newblock In {\em Proceedings of the IEEE/CVF International Conference on
  Computer Vision (ICCV)}, pages 7262--7272.

\bibitem[Touvron et~al., 2020]{touvron2020training}
Touvron, H., Cord, M., Douze, M., Massa, F., Sablayrolles, A., and J{\'e}gou,
  H. (2020).
\newblock Training data-efficient image transformers \& distillation through
  attention.
\newblock {\em arXiv:2012.12877}.

\bibitem[Vaswani et~al., 2017]{vaswani2017attention}
Vaswani, A., Shazeer, N., Parmar, N., Uszkoreit, J., Jones, L., Gomez, A.~N.,
  Kaiser, {\L}., and Polosukhin, I. (2017).
\newblock Attention is all you need.
\newblock {\em Advances in neural information processing systems}, 30.

\bibitem[Zheng et~al., 2021]{9578646}
Zheng, S., Lu, J., Zhao, H., Zhu, X., Luo, Z., Wang, Y., Fu, Y., Feng, J.,
  Xiang, T., Torr, P.~H., and Zhang, L. (2021).
\newblock Rethinking semantic segmentation from a sequence-to-sequence
  perspective with transformers.
\newblock In {\em 2021 IEEE/CVF Conference on Computer Vision and Pattern
  Recognition (CVPR)}.

\end{thebibliography}
\bibliographystyle{apalike}
%%%%%%%%%%%%%%%%%%%%%%%%%%%%%%%%%%%%%%%%%%%%%%
%%%%%%%%%%%%%%%%%%%%%%%%%%%%%%%%%%%%%%%%%%%%%%

\newpage 
\appendix
\appendixpage

\section{Model Size Used in the Neural Machine Translation Task}
\label{appendix_NMT}

\begin{table}[h!]
\caption{\small{Only the difference between the base model and the small model is listed. We use the small model size to demonstrate performance difference between sequential and parallel blocks inside the transformer layers.}}
\label{tab:model_size_NMT}
\vskip 0.1in
% \begin{small}
\centering
\begin{tabular}{c | c | c  }
    \toprule[1.25pt]
    & Base & Small \\
    \midrule[1pt]
    \Verb#encoder_embed_dim# & 512 & 128 \\
    \Verb#encoder_ffn_embed_dim# & 2048 & 512 \\
    \Verb#encoder_attention_heads# & 8 & 2 \\
    \Verb#decoder_attention_heads# & 8 & 2 \\
    % Model & \Verb#encoder_embed_dim# & \Verb#encoder_ffn_embed_dim# & \Verb#encoder_attention_heads# & \Verb#decoder_attention_heads#   \\
    % \midrule[1pt]
    % base  & 512 & 2048 & 8 & 8\\
    % small & 128 & 512 & 2 & 2\\
    \bottomrule[1.25pt]
\end{tabular}
% \end{small}
\end{table}

\section{Model Size Used in the Neural Language Modeling Task}
\label{appendix_NLM}

\begin{table}[h!]
\caption{\small{Only the difference between the base model and the small model is listed. We use the small model size to demonstrate performance difference between sequential and parallel blocks inside the transformer layers.}}
\label{tab:model_size_NLM}
\vskip 0.1in
% \begin{small}
\centering
\begin{tabular}{c | c | c  }
    \toprule[1.25pt]
     &Base & Small \\
    \midrule[1pt]
    % \Verb#decoder_layers# & 6 & 6 \\
    \Verb#decoder_embed_dim# & 512 & 128 \\
    \Verb#decoder_ffn_embed_dim# & 2048 & 512 \\
    \Verb#decoder_attention_heads# & 8 & 2 \\
    % Model & \Verb#encoder_embed_dim# & \Verb#encoder_ffn_embed_dim# & \Verb#encoder_attention_heads# & \Verb#decoder_attention_heads#   \\
    % \midrule[1pt]
    % base  & 512 & 2048 & 8 & 8\\
    \bottomrule[1.25pt]
\end{tabular}
% \end{small}
\end{table}
\end{document}